\documentclass[letter,12pt]{article}
\usepackage[letterpaper,margin=1in]{geometry}
\usepackage{setspace} 
\usepackage{authblk}
\usepackage[numbers]{natbib}
\usepackage{url}
\usepackage{graphicx}
\graphicspath{{figures/}}
\usepackage{hyperref}
\hypersetup{
    colorlinks,
    citecolor=black,
    filecolor=black,
    linkcolor=black,
    urlcolor=black
}

\renewcommand\Affilfont{\fontsize{11}{12}\itshape}
\makeatletter
\renewcommand\AB@affilsepx{ \protect\Affilfont \\}
\makeatother

\title{Cognitive science as a source of forward and inverse models of human decisions for robotics and control}

\author[1]{Mark K. Ho}
\author[1,2]{Thomas L. Griffiths}
\affil[1]{Princeton University, Department of Computer Science, Princeton, NJ, USA}
\affil[2]{Princeton University, Department of Psychology, Princeton, NJ, USA}
\date{}

\begin{document}

\maketitle

\begin{abstract}
Those designing autonomous systems that interact with humans will invariably face questions about how humans think and make decisions. Fortunately, computational cognitive science offers insight into human decision-making using tools that will be familiar to those with backgrounds in optimization and control (e.g., probability theory, statistical machine learning, and reinforcement learning). Here, we review some of this work, focusing on how cognitive science can provide forward models of human decision-making and inverse models of how humans think about others’ decision-making. We highlight relevant recent developments, including approaches that synthesize blackbox and theory-driven modeling, accounts that recast heuristics and biases as forms of bounded optimality, and models that characterize human theory of mind and communication in decision-theoretic terms. In doing so, we aim to provide readers with a glimpse of the range of frameworks, methodologies, and actionable insights that lie at the intersection of cognitive science and control research.
\end{abstract}


\tableofcontents

\section{Introduction}

As robots and other automated systems are beginning to become more integrated into human lives, engineers face a new problem: designing these systems to effectively and safely interact with people. Part of the challenge is that humans are themselves autonomous agents, making decisions and acting in ways that introduce potentially unpredictable dynamics into the environment. Even more challenging, humans change their behavior in response to the actions of the system they are interacting with, meaning that the engineer has to consider not just how to predict and interpret human behavior, but how the behavior of the system that  they are designing might be predicted and interpreted by humans in turn.

As cognitive scientists, we have had many enjoyable conversations with engineers about how to solve these problems. Typically these conversations begin with an email or a knock on the door requesting the most up-to-date model of human behavior in a format that can be easily integrated into a control-theoretic framework. We disappoint our colleagues by telling them that unfortunately no such model exists, but then excite them with how much progress has been made towards this goal and all of the research possibilities that this entails. Our intent in this article is to offer our readers a chance to follow the same emotional trajectory, highlighting the ways in which we think contemporary cognitive science can provide tools that may be useful to engineers designing systems that interact with humans and identifying some of the exciting possibilities for future research in this area.
 
\begin{figure}[!ht]
    \centering
    \includegraphics[width=.8\textwidth]{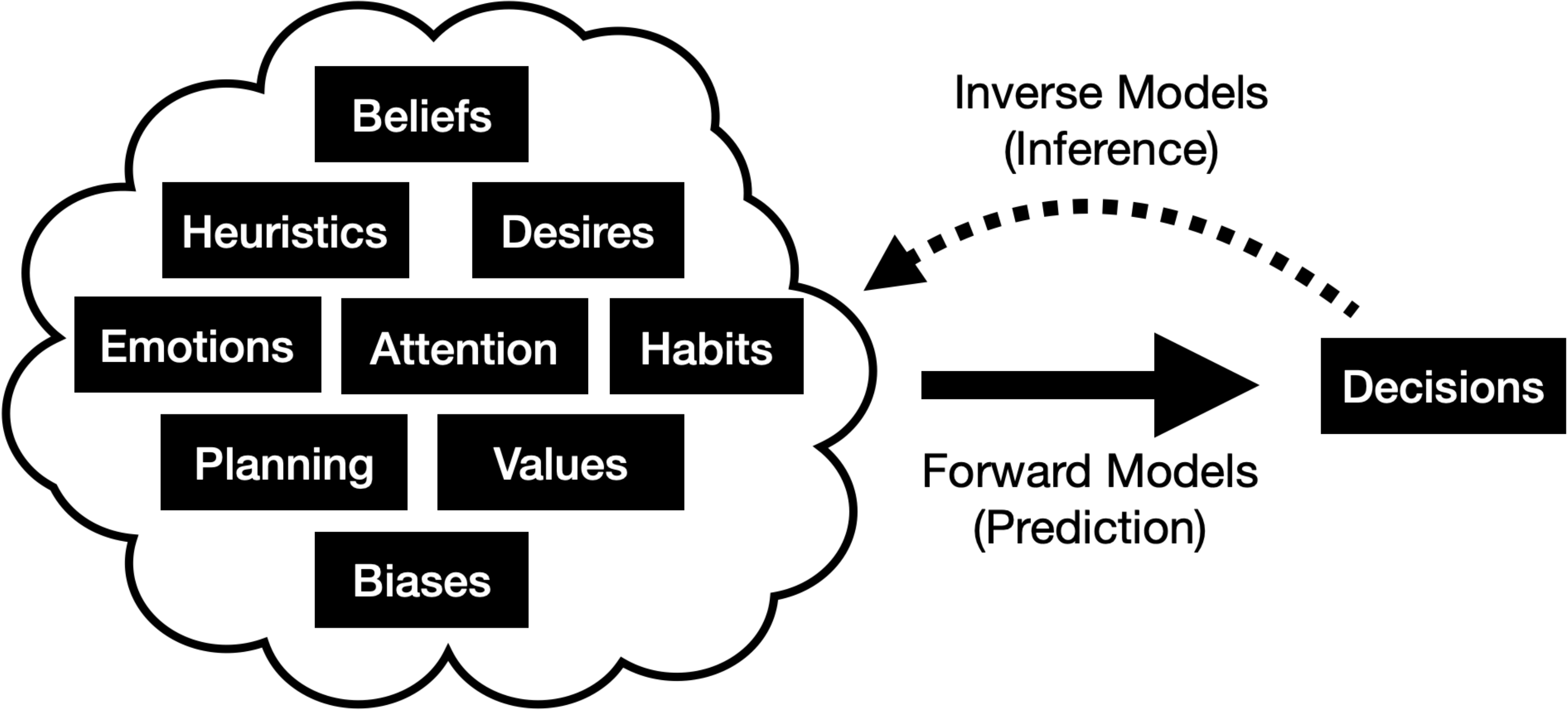}
    \caption{Forward and inverse models from cognitive science. In this paper, we review work in cognitive science on forward models of how humans make decisions and inverse models that humans use to reason about other agents. Cognitive scientists are increasingly using computational tools such as probability theory, reinforcement learning, and statistical machine learning to characterize forward and inverse models of human decision-making. This provides opportunities for cross-talk and collaboration between cognitive science and control research.} 
    \label{fig:forward-inverse}
\end{figure}

Speaking broadly,  computational models developed by cognitive scientists offer solutions to at least two of the problems that engineers face (Figure~\ref{fig:forward-inverse}). First, they provide {\em forward models} that can be used to generate predictions about human behavior based on assumptions about the beliefs, goals, and desires of human agents. These models can be useful for anticipating what a person will do in a given situation and hence provide a way to enrich modeling of the environment in which an automated system operates.  These forward models can also be used as an ingredient in {\em inverse models},  which infer the beliefs, goals, and desires of humans based on their actions -- a necessary step for any agents that seek to coordinate their behavior with humans, collaborate effectively, provide assistance, or cooperate as they pursue common goals.
 
In addition to this, cognitive science also offers insight into the way that humans solve exactly this inverse problem. People routinely make inferences about the beliefs, goals, and desires of other people, a process that has been extensively studied by psychologists \cite{premack1978does,tomasello_understanding_2005,malle2008fundamental} and is increasingly captured in computational models \cite{baker2009action,lucas2014child,jara2016naive}. These models are useful both as a source of insight for engineers seeking to re-create this capacity to draw inferences from the actions of others, but also as a tool for anticipating how the actions of an automated system will be interpreted by a human \cite{scassellati2002theory,breazeal2003toward}. Research in human-robot interaction has begun to make use of these ideas, designing systems that act in a way that is more legible to humans \cite{dragan2013legibility,ho2016showing,fisac2020pragmatic}, a line of work that we will also review.
 
In considering these two ways that computational models of cognition can be used by engineers designing automated systems -- as both forward and inverse models -- we will also highlight the ways in which recent work in computational cognitive science has emphasized formalisms that will be very familiar to researchers coming from a background of optimization and control. Cognitive scientists increasingly use ideas from probability theory, statistical machine learning, and reinforcement learning in specifying models of human cognition \cite{sun2008cambridge}. This creates an opportunity to develop a common language for describing the behavior of both humans and machines, and supports easier integration of insights from cognitive science into control.
  
Given the vast scope of human behavior, is necessary to limit our review to a specific subdomain of human activity. To that end, we will focus on models of human decision-making, broadly construed. The decisions people make reveal their preferences and determine their actions, key to the design of interactive systems. They also provide a rich territory for researchers, with formal models of human decision-making going back almost 300 years \cite{bernoulli22originally}. Our goal is to summarize the current state of the art in predicting and interpreting human decisions in a form that is immediately actionable by designers of automated systems.
  
The remainder of the paper is split into two parts. In the first part we focus on forward models, considering the criteria for useful computational models of human decision-making and summarizing recent research that aims to satisfy these criteria. We then turn to inverse models, describing the problem of inverse inference, summarizing the key ideas from the psychological  literature, explaining how this has been translated into formal models, and highlighting some of the ways in which these ideas have been applied within robotics. We close with a brief discussion of some of the remaining open questions in these areas and possibilities for future research.

\section{Forward models of human decision-making}

\begin{figure}
    \centering
    \includegraphics[width=\textwidth]{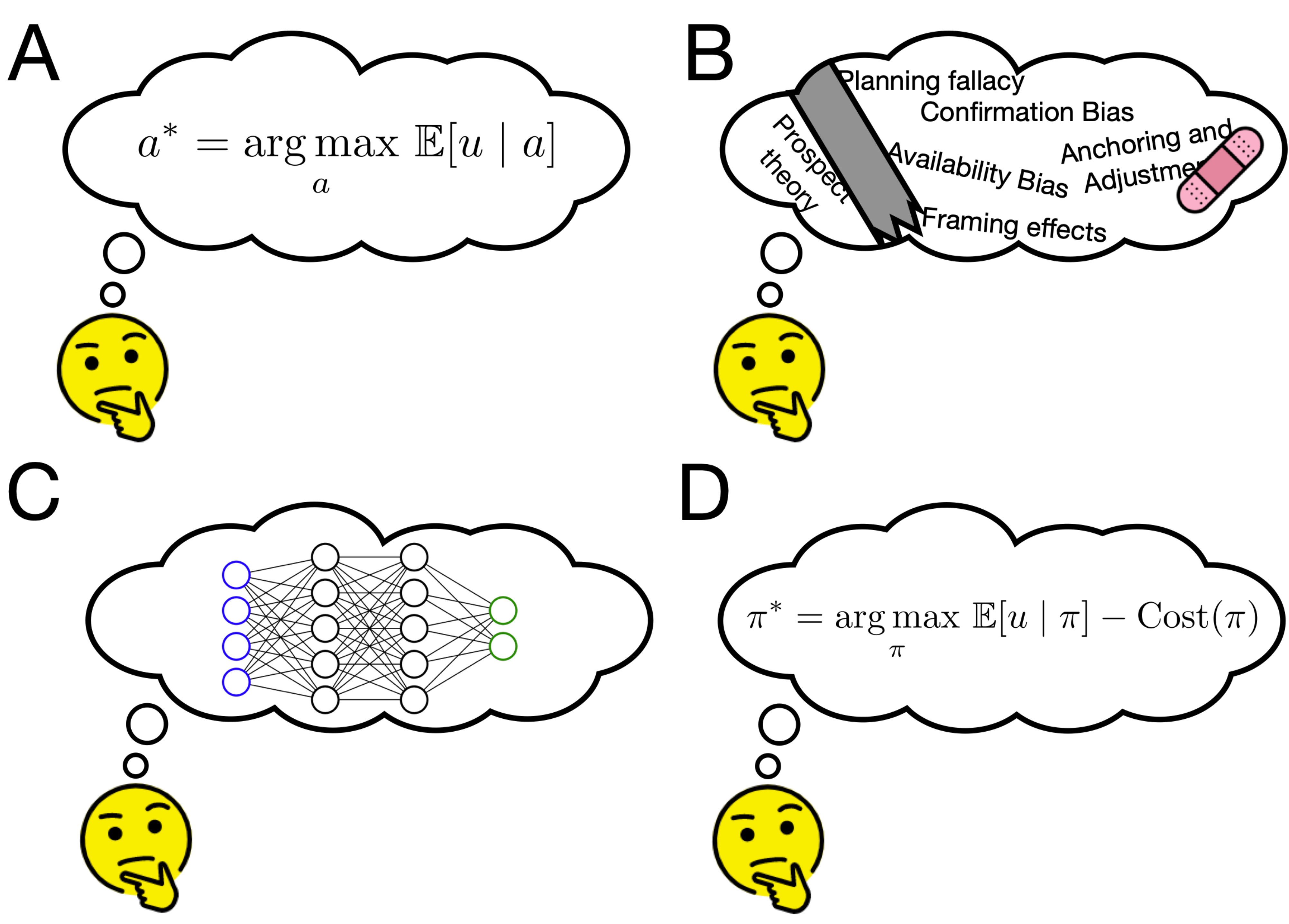}
    \caption{{\bf Four approaches to studying human decision-making} (A) Early formal theories of decision-making assumed humans were \emph{expected utility maximizers}. (B) The \emph{heuristics and biases} research program initiated in the 1970's by Tversky and Kahneman~\cite{tversky1974judgment} demonstrated that people systematically violate basic predictions of expected utility theory. This resulted in a focus on the heuristics that people tend to use in specific situations as opposed to general theories. (C) Decision-making models represented by neural networks, combined with large data sets of human choice behavior and informed by psychological theory~(e.g., \cite{peterson2021,agrawal2020scaling}), provide a way to predict about human decisions (see section~\ref{sec:pred}). (D) The theoretical framework of \emph{resource rationality}~\cite{griffiths2015rational,lieder2020resource} aims to provide a general formal theory that accounts for people's heuristics and biases. Specifically, resource rationality proposes that human decision-making reflects expected utility maximization subject to computational costs and cognitive limitations (see section~\ref{sec:rr}).} 
    \label{fig:psych}
\end{figure}

For a theory of human decision-making to be useful to an engineer designing a system that has human behavior as a component, that theory should have two properties: it should be {\em generalizable}, meaning that it can be applied in any context in which the engineer needs to be able to make predictions about how people will act, and it should be {\em accurate},  producing good predictions about human behavior in that context. The development of theories of human decision-making has historically tended to alternate between these criteria, making progress on one at the cost of the other (Figure~\ref{fig:psych}).
 
The earliest formal theories of human decision-making made  the strong assumption that humans are rational, in the sense of pursuing actions that are in their self-interest and in compliance with axioms that can be widely agreed to characterize rational behavior \cite{von1944theory,savage1972foundations}. The impressive result of this investigation is that the preferences of a rational agent can be characterized by a utility function that assigns a numerical value to each possible outcome, and that when faced with decisions that involve uncertainty that agent should pursue the option that has highest expected utility.
 
This theory -- which we will refer to as expected utility theory -- fulfills the goal of generalizability. In order to predict the actions that a rational agent will take in a new environment, it is necessary only to identify the utility assigned to different outcomes -- the decisions that agent will take can then be derived directly from these quantities. The tools that are used for deriving this behavior are exactly the tools that are used in optimization and control, as we typically seek to define agents that are rational. As a consequence, human rationality is a common assumption  in interactive systems, albeit with some allowance for stochasticity (e.g., \cite{ziebart2008maximum}). It is also a common assumption in the kind of choice models that are widely used in econometrics (e.g., \cite{mcfadden1973conditional}).
 
The only problem with assuming that humans are rational is that this assumption turns out to be false. Starting in the 1970s, psychologists (led by Daniel Kahneman and Amos Tversky) began to document the ways in which people's decisions violate the axioms that are assumed in rational models \cite{tversky1974judgment}. This led to a swing towards a more qualitative psychology of human decision-making, in which the emphasis was placed on  identifying all of the heuristic shortcuts that people seem to use when making decisions, and the behavioral biases that result. The outcome of this process is a long list of the things that people do wrong in specific situations. This is something that might potentially increase the accuracy of our models, but since these behaviors are specific to particular scenarios and it is hard to know which heuristic might dominate in a new setting, this accuracy comes at the cost of generalizability.
  
This qualitative approach to understanding human decision-making was complemented by efforts to formalize the cognitive processes that people engage in when making decisions. Kahneman and Tversky developed  \emph{prospect theory}, which extends expected utility theory by allowing different functions characterizing the subjective value of gains and losses and recognizing that the probabilities of events may also be subjectively transformed \cite{kahneman1979prospect}. Subsequent work has introduced further nuances to this theory (e.g., \cite{tversky1992advances}), together with hypotheses for how to formalize ideas about different heuristics that people might follow (e.g., \cite{gigerenzer1999simple}) as well as other cognitive factors such as the salience of different options (e.g., \cite{bordalo2012salience}).
  
Recent work in psychology and neuroscience has drilled down even further into the cognitive processes that might account for these behaviors. One prominent line of work focuses on the idea that people make decisions by accumulating evidence that one option is better than the alternatives \cite{ratcliff2016diffusion}.  There is ongoing debate about the precise mechanisms by which such a process could operate (e.g., \cite{tsetsos2012using}), but these accumulator models have also received support from results in neuroscience that seemed to show areas in the brain that engage in evidence accumulation (e.g., \cite{kiani2009representation}, but see \cite{latimer2015single}).
   
For the engineer, the precise cognitive and neural mechanisms underlying people's decisions might matter less than what those decisions actually are and how good predictions about human decision-making can be generated in other contexts -- our two criteria of accuracy and generalizability. To this end, we are going to focus on two recent developments that increase the potential for models of human decision-making to be used effectively as forward models in control settings. The first is the potential to use ideas from machine learning, combined with the availability of large data sets on human behavior, to develop more accurate models of human decision-making. The second is recent efforts to revisit the notion of rationality, with the goal of obtaining a theoretical framework that has the same generalizability as expected utility theory while incorporating what we know about human cognitive limitations in a way that supports greater accuracy.

\subsection{Developing more accurate models of human decisions using machine learning} \label{sec:pred}

The recent success of machine learning in many domains raises the possibility that such systems may be able to better predict human decisions than the theories of choice developed by psychologists and economists. For the engineer seeking a forward model of human decision-making, it may be tempting to collect a data set of human decisions and train an off-the-shelf machine learning method to predict people's behavior. This possibility has been explored extensively over the last decade, showing that machine learning has a great deal of promise in this area, but also that performance of these systems can be significantly improved by the injection of some psychological insight.

A first extensive comparison of psychological models against machine learning systems for predicting human decisions occurred in the 2015 Choice Prediction Competition \cite{erev2017_psych_review_beast}. The competition employed a standard \emph{risky choice} paradigm that has been used extensively to study human decisions, informing the development of many of the models summarized above. In this task, participants make a choice between two gambles. In each gamble different outcomes -- here corresponding to actual monetary gains and losses – occur with different probabilities. The pairs of gambles can be described by an 11 dimensional vector that summarizes the payoffs and their probabilities. The task is to map this 11 dimensional vector to a probability of choosing one gamble over the other, with the goal of getting this probability as close as possible to the choice probabilities of a group of human participants. In the competition, the choice probabilities for 90 such pairs of  gambles were provided, and the goal was to predict the corresponding probabilities for a held-out test set.

The results of the 2015 Choice Prediction Competition showed that psychological models -- that is, those developed by psychologists and economists -- tended to outperform off-the-shelf machine learning methods. The best-performing model instantiated a set of heuristics that had been identified in the psychological literature. Subsequent work showed that machine learning methods could improve on this performance, but only when provided with features that were motivated by psychological theory \cite{noti2016behavior,plonsky2017psychological}.

To machine learning practitioners, these results may not come as a big surprise. This prediction problem has a relatively large number of features compared to the amount of available data (90 pairs of gambles).  The success of psychological models can be interpreted as an instance of the  bias-variance trade-off \cite{geman1992neural}, with the small amounts of data involved meaning that models with carefully crafted inductive biases are most likely be successful. However, the other side of that trade-off is the expectation that as the amount of data increases we should expect to see improved performance from machine learning models with weaker inductive biases. 
 
Consistent with this hypothesis, applications of machine learning to predicting other kinds of human decisions have shown greater success.  With more instances of more constrained problems, machine learning methods can outperform psychological models, and have even been suggested as offering an upper bound on the amount of variance we can expect to account for \cite{fudenberg2019measuring,peysakhovich2017using}.  Indeed, in a subsequent Choice Production Competition where models were trained on 210 pairs of gambles, relatively generic machine learning models were able to outperform psychological theories \cite{plonsky2019_cpc18}.

Based on this insight, recent work has collected and analyzed a risky choice dataset that involves orders of magnitude more problems than the original Choice Prediction Competition \cite{bourgin2019cognitive,peterson2021}. In this data set, human participants made decisions for over 10,000 pairs of  gambles. The size of the data set makes it possible to systematically evaluate existing models of choice, and to use machine learning to exhaustively explore the space of possible theories. Different models of choice can be expressed in terms of constraints on the functional form of a predictive model. For example, under the expected etility theory we can take the probability that people choose a gamble to be proportional to $\exp \{\sum_i p_i u(x_i)\}$  where $p_i$ is the probability of the outcome $x_i$ and $u(\cdot)$ is a utility function. By taking an arbitrary differentiable form for this utility function -- such as an artificial neural network -- we can employ standard tools for automatic differentiation to use gradient descent to optimize the form of this function against human data.  This approach generalizes to other psychological theories. For example, prospect theory corresponds to assuming the choice probability is proportional to $\exp \{\sum_i \pi(p_i) u(x_i)\}$ where $\pi(\cdot)$ is a probability weighting function.

\begin{figure}
    \centering
    \includegraphics[width=\textwidth]{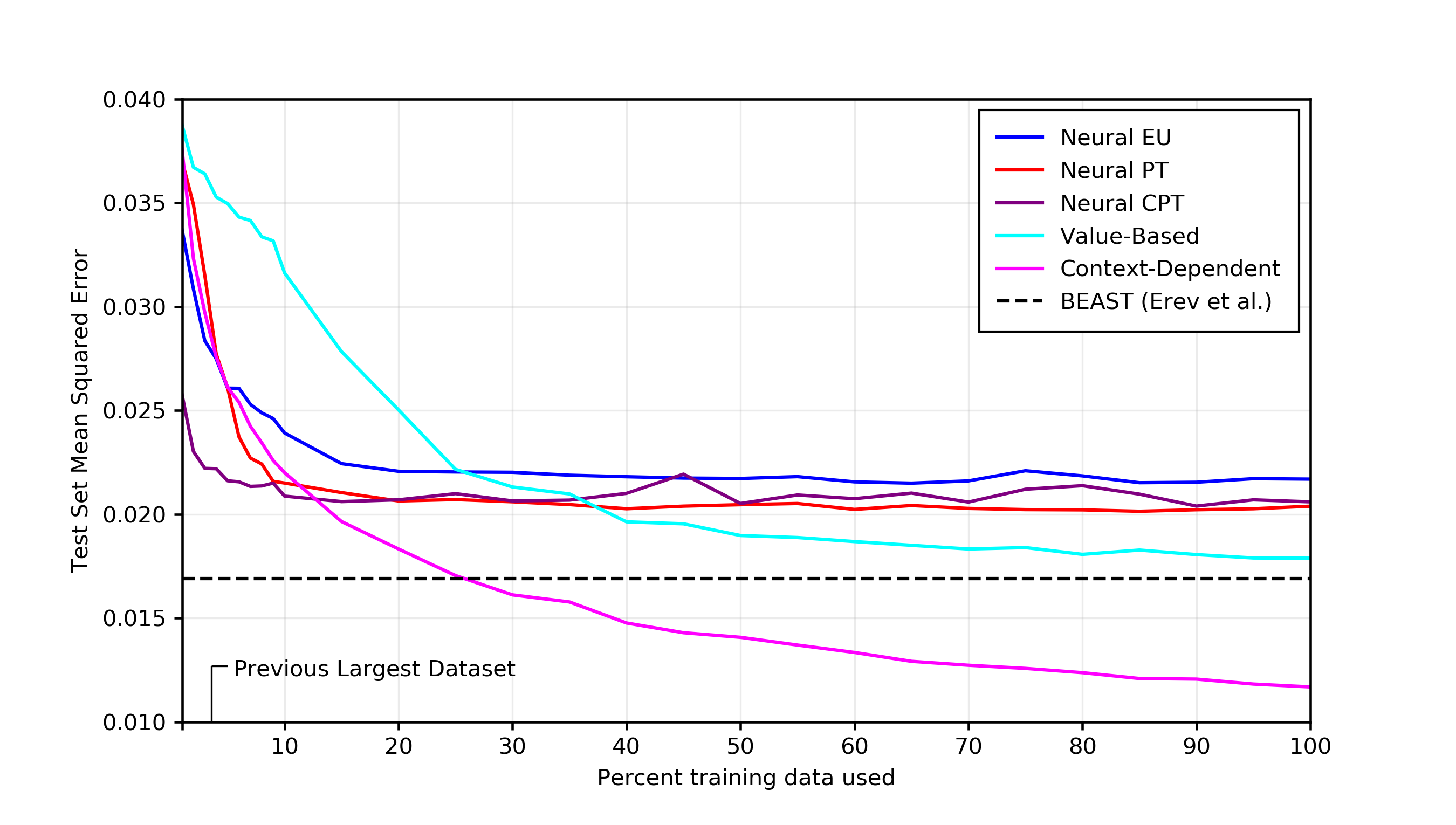}
    \caption{Performance of machine learning models protecting people's decisions in a risky choice task (data from \cite{peterson2021}).  Neural networks constrained to a functional form consistent with classic theories of decision-making such as expected utility (EU), prospect theory (PT), and cumulative prospect theory (CPT) are compared against networks that directly estimate the value of a gamble from its features (Value-Based) or directly predict people's choices based on the features of both gambles (Context- Dependent). The vertical axis shows means squared error in predicting the probability with which people choose a particular gamble, the horizontal axis shows the percent of The training data (approximately 10,000 pairs of gambles) that was used. The previous largest data set for risky choice \cite{plonsky2019_cpc18} is shown.  Given enough data, all neural network models outperform the best models in their class proposed by human psychologists and economists, but this requires orders of magnitude more data than have previously been collected. For comparison, the dotted line shows the performance of the Best Estimate And Sampling Tools (BEAST) model\cite{erev2017_psych_review_beast} that won the 2015 Choice Prediction Competition.} 
    \label{fig:peterson}
\end{figure}

Peterson et al. \cite{peterson2021} used this approach to identify the optimal functional form for various classic theories of choice, and also evaluated unconstrained artificial neural networks for predicting people's decisions (see Figure \ref{fig:peterson}). The results showed that when using the entire data set of different pairs of gambles, an unconstrained neural network systematically outperformed all existing psychological theories. However, they also showed that equivalent performance could be obtained by defining a model based on a mixture of these classic  theories, and that this model achieved a high level of predictive performance far faster than an unconstrained neural network. The results of this analysis suggest that, given enough data, we can obtain better forward models of human decisions using machine learning, but that these models are likely to be enhanced by drawing on psychological theory when possible.

A further challenge of using off-the-shelf machine learning methods when developing forward models of human decision-making is that these methods often result in models that are  uninterpretable. Previous work in this area has relied on post hoc analysis of models to identify features that are psychologically interpretable (e.g., \cite{peysakhovich2017using}).  An alternative approach was recently outlined by Agrawal et al. \cite{agrawal2020scaling}, in which an off-the-shelf machine learning model is used to critique a more interpretable model until that interpretable model yields similar performance. This approach was applied to a large data set of human decisions that is likely to be of interest to researchers working on autonomous systems: the Moral Machine project \cite{awad2018moral}. This data set consists of more than 10 million human decisions about what an autonomous vehicle should do when faced with an inevitable collision, where the only available choice is about which group of pedestrians the vehicle will collide with (a version of the classic trolley problem \cite{thomson1976killing}). Having an uninterpretable model that predicts these choices is not particularly useful for designing autonomous systems, but Agrawal et al. showed that their approach can be used to identify the features that a predictive model had discovered, making those features explicit in a way that is likely to be useful when deciding how to design and regulate autonomous vehicles.

\subsection{Developing more generalizable models via resource rationality}
\label{sec:rr}

 Despite the promise of machine learning to improve the accuracy of a models of human decisions, generalizability is still likely to be a challenge. Machine learning systems are typically trained in a specific domain, and  can face difficulty when applied in another related domain. Making this kind of generalization requires extracting the causal principles that underlie  people's decisions. In this section we consider an approach that has the potential to do just that.
 
 The classical notion of rationality was our prime example of a theory that satisfies the goal of generalizability -- for any new situation, it is possible to derive predictions about behavior. However, this classical theory falls short not just because it fails to empirically capture aspects of how people make decisions, but because it represents an unrealistic ideal for any intelligent system with finite computational resources.  This is a long-standing idea, going all the way back to the classic work of Herbert Simon on \emph{bounded rationality} \cite{simon1955}. However,  this idea has  recently begun to receive a more comprehensive mathematical and empirical treatment.
 
 The classical theory of rational action via maximizing expected utility doesn't take into account the computational cost of selecting that action. As a consequence, it's easy to imagine an agent trying to follow the prescriptions of this theory ending up paralyzed as it tries to compute all of the possible outcomes and their probabilities. To address this, researchers in the artificial intelligence literature have sought a more realistic criterion for rational action for agents with finite computational resources. The outcome of this investigation is the theory of bounded optimality, which focuses not on the optimal action that an agent should take but rather on the optimal algorithm an agent should follow in order to select that action \cite{Horvitz1987,russell91}. This theory explicitly trades off the expected utility of  finally taking an action with the computational cost that's involved in getting to that point.
 
 For cognitive scientists, bounded optimality offers a way to theorize about the optimal cognitive processes that intelligent agent should engage in when trying to make a decision \cite{gershman2015computational,lewis2014computational}. As it puts an emphasis on rational use of the cognitive  resources an agent is able to apply, this approach has been referred to as \emph{resource rational analysis} \cite{griffiths2015rational,lieder2020resource}. Considering how an agent should rationally deploy its cognitive resources provides a way to explain why people may choose to adopt particular heuristics -- even if those heuristics result in systematic biases -- and to make generalizable predictions about the kinds of cognitive strategies that we expect people to engage in.

 Recent research has recast some of the classic heuristics discovered by psychologists from the perspective of  the rational use of cognitive resources. For example, focusing on extreme events when considering the outcomes of a decision is a strategy that can minimize the variance of estimates of expected utility based on small samples, even though it introduces a bias to those estimates \cite{lieder2018overrepresentation}. Thinking in these terms allows us to potentially begin to reconcile the various heuristics and biases identified by psychologists into a broader mathematical theory.  

 To the engineer, resource rationality offers the potential to make better predictions about human behavior in a way that incorporates realistic assumptions about the cognitive limitations of human agents. Generalizability results from the fact that  deriving an optimal resource-rational strategy can be formulated as a sequential decision problem. An agent trying to make a decision is going to execute the sequence of computations that provide information about the possible outcomes of their actions, at some point selecting an action to perform based on this  information. The sequential decision problem here corresponds to the choice of that sequence of computations: we can construe each computation as a kind of mental action, ending with the decision that we have done enough computation and are ready to act as the end of the sequence.

Expressed in these terms, it is possible to see that we can use familiar tools such as Markov decision processes to formalize the internal decision-making we do about how to deploy our cognitive resources.  Solving the resulting MDPs (referred to as \emph{meta-level MDPs} \cite{Hay2012}) provides a way to derive predictions about behavior.  Crucially, this approach carries with it the same generality as the classic theory of rational action. It simply moves rationality up the level of the choice of how to deploy cognitive resources, and requires us to be explicit about what those resources might be. 

To provide a concrete example, one recent paper \cite{callaway2021fixation} used this approach to examine how can model attention allocation in a simple decision-making task. In this task people are presented with three objects -- in this case snack foods -- and asked to decide which they prefer. While they are doing this, their gaze it recorded using an eye tracker.  People show a consistent pattern of behavior in this task. For example, they  spend more time looking at items that they assign a higher subjective value.  These patterns of behavior can be explained by assuming that people are trying to estimate the subjective value of each item, and that each moment they spend looking at an option provides a sample from a Gaussian distribution centered on that value. The problem of deciding whether to sample can then be formulated as an MDP, and the resulting policy generates predictions about which objects people look at and in what sequence. We spend more time looking at items with higher subjective value because those are the items that are most relevant to our ultimate decision. 

Describing cognitive processes in terms of the solution to Markov decision processes has the virtue of characterizing human cognition using formal tools that are likely to be familiar to those working in control theory. Other work on resource rationality has likewise employed formalisms that will be familiar to engineers. For example, one line of work focuses on the information-theoretic costs of maintaining mental representations at a given degree of precision \cite{ortega2011information,bhui2018decision}.  This approach also has connections with work in economics that explains apparently irrational aspects of human choice in terms of \emph{rational inattention},  where information-theoretic costs are assumed to apply to the precision of the signal an agent uses to inform a decision \cite{sims2003implications,gershman2020rationally}. 

Resource rationality offers a new set of tools for capturing human behavior with greater precision in a way that is compatible with standard modeling techniques used in robotics and control.  Being able to make generalizable predictions about how long it will take people to make a decision, what information they going to seek when making that decision, and what kinds of information are likely not to include when making a decision are all things that can facilitate the design of human-machine interfaces. Furthermore, it is possible to use this kind of approach to engage with the question of how to improve human decision-making: if we assume that people are rational but resource-limited, we can think about how assistive robots might modify the environments in which humans are making decisions to allow them to make better use of those resources. 
 
\section{Inverse models of human decision-making}

While forward models can help us to make predictions about what people will do given their preferences, this is typically not  all we need to know in order to design systems that are able to interact effectively with humans. In an ideal world, autonomous systems would effectively help fulfill people's needs and desires, which can change from person to person and situation to situation.  We need a way to infer those needs and desires from people's behavior -- inverse models. Furthermore, people are adaptive; they often change their behavior in response to a system based on their best guess as to how it functions and may even expect the system to do the same. So it is not enough to be able to make inferences about people's needs and desires, we also need to anticipate the inferences that people will be making in turn.

Developing systems that can solve these problems is a daunting challenge, but fortunately, we can take inspiration from existing systems that must regularly interact with humans: other humans. And while nearly everyone has had to deal with other humans, cognitive science offers an extensive, systematic understanding of how we effectively solve the problem of understanding and interacting with others in our everyday lives.

In this section, we turn our focus towards what cognitive science has to say about people's inverse models of cognition and action. One of the most remarkable capacities that humans have is the ability to understand the hidden mental states that give rise to other people's observable behavior~\cite{gergely2003teleological,malle2008fundamental}. This ability is often referred to as \emph{theory of mind}, and cognitive scientists have studied it in adults~\cite{malle2008fundamental}, children~\cite{flavell2004theory}, infants~\cite{gergely1995taking}, and even other species~\cite{premack1978does}. Theory of mind has played a key role in our evolution as a social species capable of large-scale culture, coordination, and cooperation, and its development within the first year of life is a major milestone that enables us to comprehend and participate in the social world~\cite{tomasello_understanding_2005}.

One exciting development in the recent study of theory of mind has been the application of ideas from economics, artificial intelligence, and control theory to characterizing mental state inference in computational terms. These approaches are reminiscent of methods familiar to engineers such as imitation learning, inverse reinforcement learning, and apprenticeship learning~\cite{abbeel_apprenticeship_2004,argall2009survey,arora2021survey}. However, in modeling the varieties of human social inference and interaction, they depart from and extend these ideas in numerous ways. This presents an exciting opportunity for collaboration between the cognitive sciences and engineering by providing new perspectives on inverse decision-making but in a shared conceptual and technical framework.

Here, we will focus on two lines of research on inverse models at the intersection of artificial intelligence and cognitive science. The first is work on cataloguing and systematizing the \emph{conceptual primitives} involved in mental state inference---e.g., how people reason about mental entities like beliefs, desires, intentions, emotions, etc. The second is on inverse models in the context of teaching and communication, which are among the most basic types of social interactions that also expose the complexity of how humans use theory of mind productively. Along the way, we will discuss cases in which ideas from cognitive science have already been applied to the design of automated systems, limitations of existing approaches, and the possibilities for future research and applications.

\subsection{Identifying the building blocks of theory of mind}

Put in simple computational terms, theory of mind is an inference problem. That is, given limited observations of a process (e.g., a person's behavior), the task is to identify the hidden variables that produced those observations (e.g., the person's thoughts, desires, or feelings). Of course, this requires not only having concepts like \emph{thoughts} and \emph{desires} but an understanding of how these elements combine to produce behavior. This can be understood in rough analogy to another machine learning problem: parsing natural language. For example, inferring the \emph{parse tree} of a particular sentence is jointly constrained by knowledge of \emph{primitive types} of words (e.g., nouns, verbs, prepositions) and a \emph{grammar} of how words tend to be combined. Recent models of theory of mind can be understood in terms of this linguistic metaphor: To explain how humans \emph{parse} the behavior of other agents, we must understand the mental state primitives and mental state grammars that dictate how they are combined.

Incidentally, we have already covered one possible theory of how people parse behavior: expected utility theory~\cite{von1944theory,savage1972foundations}. Taken as a generative model of people's intentional action, expected utility theory posits that others have beliefs about the state of the world (e.g., the belief that there is a burger joint down the street) and desires that certain states of the world are realized (e.g., the desire to eat a burger for lunch) and that people act rationally to realize their desires given their beliefs (e.g., the act of walking down the street to the burger joint). As an account of theory of mind, \emph{inverse expected utility theory} makes several generalizable predictions that have been confirmed with human experiments. For example, adults, children, and infants can reason about how others integrate information about goals and action costs~\cite{baker2009action,jara2015children,liu2017ten}, features of different choices~\cite{jern2017people}, the statistics of the environment~\cite{lucas2014child}, and limited perception of the environment~\cite{baker2017rational}. Findings such as these have led to the proposal that human common sense psychology consists of a \emph{na\"{i}ve utility calculus} where we abstractly reason about other decision-makers as utility-maximizing agents~\cite{jara2016naive}.

At a broad level, the formal tools used characterize inverse utility theory will be familiar to those from a control background as they build on standard formalisms like MDPs, POMDPs, and inverse reinforcement learning~\cite{jara2019theory}. However, there are interesting differences between how they are applied as models of human inference versus their typical engineering applications. For example, cognitive models tend to assume that people have highly structured representations of others' mental states (e.g., discrete objects), whereas in engineering applications the state representations are less structured (e.g., weights on a vector of continuous features~\cite{arora2021survey}). This reflects the fact that cognitive scientists aim to explain how people can rapidly draw inferences based on only a few observations and a large base of background knowledge, while engineers are often trying to analyze large data sets of expert trajectories with minimal fine-tuning. As a result, cognitive scientists tend to use Bayesian methods while engineers are likely to be more familiar with methods designed to scale to large data sets. An important direction for future work is developing methods that cut across these different research agendas and can replicate the sophistication of human theory of mind with tractable implementations.

Expected utility theory captures an important dimension of how humans parse others behavior in terms of beliefs, desires, and intentions. But, psychologists have also long studied other types of mental states that people reason about, such as emotions, habits, norms, rules, values, and social affinities, to name only a few~\cite{malle2008fundamental}. While the traditional frameworks of expected utility and inverse reinforcement learning have not typically focused on these kinds of mental states, cognitive scientists have made great strides in extending the formalism to study these types of representations. For example, inverse planning models have been combined with models of habits~\cite{gershman2016plans}, emotion and appraisal~\cite{ong2015affective,saxe2017formalizing,ong2019computational}, responsibility judgments~\cite{gerstenberg2018lucky}, values and norms in moral dilemmas~\cite{kleiman2015inference}, and social groups~\cite{lau2018discovering,shum2019theory}. Although do we do not typically think of robots as having these types of internal states, systems that are expected to interact with humans invariably need a basic understanding of the complete repertoire of psychological states that affect our individual and collective behavior.

The models described so far generally rely on assuming the standard formulation of rational action as optimizing a utility function defined over states of an MDP. For instance, they can express the idea of reaching a goal state while attempting to minimize costs along the way, but they cannot generally express history dependent or temporally specified constraints such as only going to a goal state only after accomplishing a subgoal. Recent work in both cognitive science and computer science have attempted to remedy this by introducing more flexible ``logics’’ to express an agent's utilities, including linear temporal logic~\cite{littman2017environment,velez2017interpreting,vazquez2018learning}, finite state machines~\cite{icarte2018using}, and simple programs composed of sub-processes~\cite{ho2018human}. These methods are powerful because they can express rational cognition and action in a rich, compositional manner that may reflect human intuitions about agency. At the same time, this expressiveness comes at the cost of more complex and costly inference, and identifying appropriate constraints and settings in which tractable inference methods can be applied is an active area of research. 

A complementary method for sidestepping strong assumptions about the structure of rational action is to try and learn a theory of mind directly from data without recourse to pre-defined structural priors (e.g., using a neural network). Rabinowitz et al.~\cite{rabinowitz2018machine} take this approach by generating a large data set of behaviors from synthetic agents with different goals, utilities, and perceptual abilities, and then using meta-learning to train networks with no prior conception of theory of mind or rationality to predict features of future behavior. Their networks were able to acquire a low dimensional embedding of behaviors and agent types capable of recreating several qualitative findings associated with theory of mind, such as inference about goals, costs, perceptions, and, to a certain extent, false beliefs. This work is an important demonstration of how relatively standard machine-learning methods can learn theory of mind-like representations given enough data and computation. However, studies by Nematzadeh et al.~\cite{nematzadeh2018evaluating} have indicated that standard neural networks are limited in their capacity to explicitly represent false beliefs as they do not distinguish between appearance and reality, which is considered by psychologists to be a defining feature of theory of mind~\cite{flavell2004theory}. This suggests that some kind of structural priors about the nature of rational action are likely needed to capture the conceptual primitives and grammar comprising human theory of mind.

\subsection{Identifying generalizable principles of communication and teaching}
The previous section focused on theory of mind as a pure inference problem, but theory of mind also influences how people act and interact. Building on ideas originally explored by philosophers~\cite{wittgenstein1953philosophical,grice_meaning_1957,sperber1986}, developmental psychologists and linguists have extensively studied the principles underlying how people use theory of mind to learn from others and communicate~\cite{clark1996,tomasello_understanding_2005,csibra2009natural}. A key idea is that when people communicate (e.g., by saying words, making gestures, providing examples, etc.), they do so with \emph{communicative goals} like modifying the receiver's mental state or future actions. The person receiving these communicative signals can then reason about these goals to flexibly and efficiently draw inferences about what the sender meant to convey. Note that this process requires both the receiver and sender to have a capacity for theory of mind and, more specifically, a capacity for \emph{recursive theory of mind}, in which one agent reasons about another agent reasoning about the original agent (and potentially up to higher levels of recursion). Computational research over the past few years has formalized and extended many of these findings within the framework of probabilistic inference and decision-making~\cite{shafto_rational_2014,goodman2016pragmatic}. Here, we provide a broad overview of these developments.

One approach to studying communication in computational terms is the Bayesian pedagogy and cooperative communication framework~\cite{Shafto2008,shafto_rational_2014,landrum2015learning}, which characterizes how a teacher who presents data and a learner who interprets data should coordinate to efficiently and successfully communicate. To illustrate this idea, suppose you wanted to teach someone the concept of the even numbers by giving them a series of examples. Some examples will be better than others, for instance, \{2, 2, 2\} is technically a sequence of even numbers, but is not very informative, whereas \{2, 4, 6\} is clearly more helpful. Additionally, if the person receiving these examples knows you are being helpful, they can draw even stronger inferences based on what they are shown. Bayesian pedagogy models formalize this intuition about helpful, \emph{informative} examples in terms of recursively defined teacher-learner equations, whose fixed points are optimal teacher-learner communication protocols. This approach has been successful in characterizing how both adults and children teach and learn concepts during pedagogical interactions~\cite{bonawitz2011double,bridgers2020young}. Additionally, recent theoretical work has established a direct correspondence between optimal transport problems and the equations that characterize teacher-learner fixed points~\cite{wang2020mathematical,shafto2021cooperative}. This opens the door for algorithmic insights to be shared between engineering disciplines such as operations research and the study of optimal cooperative communication in humans.

Cooperative communication models capture settings in which a teacher has a sole, explicit goal of being helpful and informative. However, communication is not always so clear-cut, and cognitive scientists are modeling more complex communicative situations within the \emph{Rational Speech Act} (RSA) framework~\cite{goodman2016pragmatic}. For example, people may have goals other than being informative, like being succinct or inoffensive, and they may expect others to perform joint inference over these \emph{non-communicative} goals. This process can give rise to linguistic phenomena as varied as hyperbole (``This kettle cost \$1,000!''~\cite{kao2014nonliteral}) and politeness (``Your poem wasn't bad!''~\cite{yoon2020polite}). And while it may be extreme to expect automated systems to understand ironic humor, they will likely need to recognize more mundane forms of indirect speech.

More broadly, RSA can characterize how the context surrounding communicative acts shapes their meaning. For example, the statement ``It's warm today'' has different consequences if said during the spring versus the winter (e.g., you would likely wear shorts in the first but not the second case)~\cite{tessler2017warm}. Computational cognitive models of how humans flexibly reason about the shared context has been shown to be a key part of understanding general statements about the world~\cite{tessler2019language}. Similarly, the ability to establish contexts as communicative (e.g., intentionally getting someone's attention in order to convey some information) has been shown to be an essential precursor to the types of cooperative communication interactions discussed above~\cite{csibra2010recognizing,scott2014speaking}. Formal models that combine RSA with sequential decision-making and inference about whether partners have informative goals can provide a starting point for implementing such flexible reasoning about communicative context in autonomous systems~\cite{shafto2012epistemic,ho2021communication}.

\subsection{Applying human inverse models to the design of autonomous systems}

\begin{figure}
    \centering
    \includegraphics[width=\textwidth]{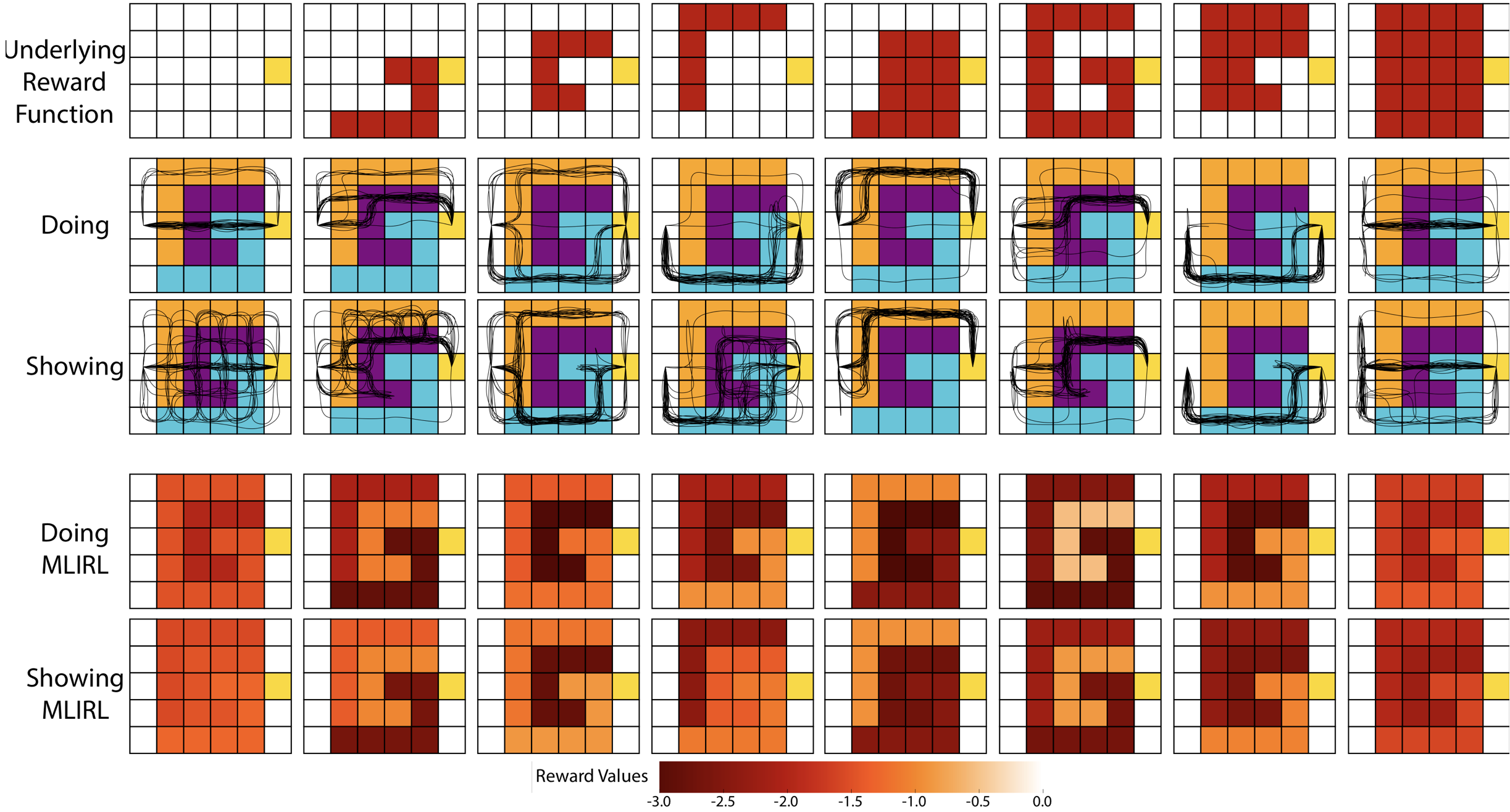}
    \caption{Learning from humans with communicative intent (data and figure from~\cite{ho2016showing}). Ho et al. recruited human participants to perform grid navigation tasks that required reaching a goal state while not losing points. Each column represents one trial. Row 1 represents the true underlying reward function where white tiles are 0 points, red tiles are -2 points and the yellow goal tile is 10 points. Participants could not directly view the reward values, but were shown colors on the grid (orange, purple, blue) and told the value of each color (e.g., ``orange and purple are safe''). Row 2 shows the visible layout of each grid and each black line represents one participant's trajectory on the task when they were only told to do the task. Row 3 shows the same grid and trajectories for participants told to do the task as well as \emph{show} the reward function to an anonymous observer. Rows 3 and 4 show the reward weights estimated by Maximum-Likelihood Inverse Reinforcement Learning (MLIRL)~\cite{macglashan2015between} when given the \emph{Doing} versus \emph{Showing} demonstrations. Agents trained by Showing demonstrations obtain better estimates of the underlying reward function than those trained by Doing demonstrations.}
    \label{fig:honeurips}
\end{figure}

Ideas from the computational cognitive science of mental state inference and communication have already begun to inform research in human-robot interaction and reinforcement learning. This showcases the potential for scaling up probabilistic models of cognition to complex, real-world domains, while also revealing novel insights about communication and teaching. For example, work in motion planning has led to robots capable of legible motion~\cite{dragan2013legibility,fisac2020pragmatic}, in which action sequences are modified to allow humans to more quickly and successfully understand a robot's goals. Related work by Ho et al.~\cite{ho2016showing} showed that when learning from human demonstrations, inverse reinforcement learning algorithms can benefit from being shown intentionally communicative expert behaviors (Figure~\ref{fig:honeurips}). These modifications by both robots and humans are directly analogous to the approach taken in the Bayesian pedagogy framework, and demonstrate how consideration of communicative goals can facilitate human-machine teaching, learning, and cooperation.

Insights from cognitive science can also inform the design of learning algorithms themselves. For example, humans readily use rewards and punishments to modify the behavior of other animals and even other humans, and there is good reason to believe that similar principles of mental state inference and communication apply to these interactions~\cite{ho2017social}. In a series of experiments with humans interacting with different learning algorithms, Ho et al.~\cite{ho_teaching_2015,ho2019people} demonstrated precisely this. Specifically, they found that people do not use rewards in a manner consistent with the standard interpretation as a quantity to directly maximize, as is typically done in reinforcement learning. Rather, they expect learners to reason about a teacher's pedagogical goals and interpret rewards as signaling information about whether an agent is ``headed in the right direction’’ during the learning process. Such findings help motivate the development of learning algorithms that interpret reward in more sophisticated ways and attempt to infer people’s teaching strategies and goals~\cite{loftin2014strategy,hadfield2017inverse}. 

Along similar lines, MacGlashan et al.~\cite{macglashan2017interactive} found that the structure of a human teacher's feedback depends on an agent’s current stage of learning--a simple form of context. In a behavioral study, the authors had human participants interact with either a completely na\"{i}ve agent or an expert agent that knew the optimal path to a goal. When the naive agent took a sub-optimal, but moderately good action, participants provided a high reward, while they gave the expert agent who should have known better a low or even negative reward for taking the same action. The logic of this strategy closely resembles that of the \emph{advantage function}, which reflects the relative value of each action in a state under the agent's current policy~\cite{baird1995residual}. This insight motivated the design of the Convergent Actor-Critic by Humans (COACH) algorithm, which treats human feedback as an advantage signal. Subsequent work has also successfully applied COACH to training deep learning agents~\cite{arumugam2019deep}.

Additionally, researchers in control and robotics have developed unifying frameworks for modeling cooperative interactions likely to be faced in human-robot applications. For instance, cooperative inverse reinforcement learning~\cite{hadfield2016cooperative} defines a general class of games from which specific cooperative strategies can be derived (e.g., legible motion, requests for information, etc.) based on a particular set up. Similarly, reward-rational learning~\cite{jeon2020} has been proposed as a framework for characterizing different types of human-robot interaction problems (e.g., learning from demonstrations, feedback, or examples) in terms of inference about human preferences that may be implicit. Combining these general computational frameworks with insights from the cognitive science of human decision-making will be an important direction for future research.

\section{Opportunities for further research}
The research we have reviewed is only a starting point for exploring potential applications of cognitive science to engineering, robotics, and control. There are a number of exciting directions based on these ideas we have covered. Here, we focus on three broad future directions.

{\bf Inverse resource rationality} As we have noted, cognitive scientists have proposed that people reason about others as expected utility maximizers. Additionally, we discussed how expected utility theory is inaccurate and how resource rationality is a promising alternative framework. This raises the obvious possibility that people understand that others have limited cognitive resources and therefore reason about them not as pure utility maximizers, but as \emph{resource-rational utility maximizers}. Importantly, progress here will depend on the continued development of plausible forward resource-rational models, especially models of planning (e.g.,~\cite{correa2020resource,ho2021control}). Nonetheless, several lines of research have already begun to explore inverse resource rationality. For example, research on perspective-taking and communication has shown that people can flexibly reason about the division of cognitive labor required for efficient communication~\cite{hawkins2021division}. Additionally, recent work has demonstrated how humans can infer preferences by jointly reasoning about the time it takes to make a decision and the decision itself~\cite{gates_callaway_ho_griffiths_2021}, while other studies have shown how people reason about sub-optimal or inconsistent planning~\cite{evans2016learning,alanqary2021modeling,berke_jara-ettinger_2021}. These ideas have begun to be incorporated into inverse reinforcement learning settings~\cite{zhi2020online}. As with models based on inverse expected utility theory, understanding people’s inverse models of resource-rational processes will be essential for how they interpret how machines make decisions and can also serve as inspiration for how machines interpret and interact with humans.

{\bf Blackbox vs. Theory-driven models for inferring intent}
In our discussion of forward and inverse models, we encountered three different examples of how machine learning tools can complement traditional psychological theory building. This included work on how humans make risky choices, work on how humans make moral decisions, and work on learning theory of mind concepts without hard-coding conceptual primitives or a principle of rationality. These approaches have illustrated how large data sets can be combined with machine learning tools to search the vast space of cognitive models in an efficient manner. However, they also illustrated some of the limitations of a purely blackbox approach and the benefits of also incorporating explicit, interpretable theories and structured prior knowledge. Thus, an important direction for future work will be developing methods that seamlessly integrate the benefits of each approach---on the one hand the scalability of blackbox methods, and on the other hand, the efficiency and interpretability of explicit psychological theories.

{\bf Schemas and mechanisms for human-robot interaction} Unlike settings involving a single agent, interactions between two or more agents are challenging to design because they are fundamentally ill defined: Each agent might have their own goals and beliefs, which means there may not exist a single ``yardstick’’ by which to measure whether the engineering problem has been solved. This has occasionally prompted researchers to ask themselves ``If multi-agent learning is the answer, what is the question?’’~\cite{shoham2007if}. 

We propose that cognitive science is uniquely positioned to provide guidance to question that multi-agent learning answers in the form of schemas and mechanisms grounded in the types of interactions that humans are adapted for. We have already encountered one example of this: The interaction of a teacher and learner engaged in cooperative communication can serve as a template for developing robots capable of legible action and value alignment~\cite{dragan2013legibility,hadfield2016cooperative}. Beyond this, there are many other types of human interactions and socio-cognitive mechanisms that we have not discussed that could inspire future research. For example, humans form joint intentions to achieve shared goals, and this underlies our ability to cooperatively solve novel problems~\cite{kleimanweiner2016coordinate}. At a broader scale, human interactions are often shaped by norms, which can be understood as shared behavioral tendencies that generalize across interactions with different agents in a population. Norms and normative cognition have been extensively studied in cognitive science and psychology, and researchers have begun to explore these processes computationally~\cite{hawkins2019emergence}. Finally, an additional benefit of designing autonomous systems around how humans actually interact is the potential for new insights into those very interactions, leading to further collaboration between the cognitive sciences and engineering disciplines.

\subsection{Conclusion}
At the moment, there is no unified model of human cognition and decision-making that engineers can draw on when designing their systems. Nonetheless, cognitive science has much to offer those designing autonomous systems that interact with humans. In particular, cognitive science has a rich trove of theories and methods for systematically studying how humans think, decide, and interact with one another, and these discoveries are increasingly being couched in formal terms that are familiar to researchers in robotics and control. Here, we have discussed several recent frameworks and methodologies, such as the synthesis of blackbox and theory-driven methods, resource-rational decision making, cooperative cognition, and rational speech act theory. We then surveyed how they have been applied to derive insights into the mechanisms and principles underlying people’s forward and inverse models of decision making. As autonomous systems continue to become more commonplace in people’s everyday lives, we expect engineers will also need to think systematically about how the humans their systems encounter will make decisions. We hope this review can provide clarity into the types of actionable insights and open questions that sit at the intersection of cognitive science and control research.

\vspace{2em}
\noindent {\bf Acknowledgements}:
This work was funded by NSF grant \#1545126, John Templeton Foundation grant \#61454, and AFOSR grant \# FA 9550-18-1-0077. Emojis in figures designed by OpenMoji, the open-source emoji and icon project. License: CC BY-SA 4.0.

\bibliographystyle{ar-style3.bst}
\bibliography{references}

\end{document}